\newif\ifarxiv
\newif\ifgecco

\arxivtrue \geccofalse

\ifarxiv
\documentclass[english,a4paper,11pt]{article}
\usepackage{fullpage}
\else
\documentclass
[
    sigconf = true,     
    review = false,      
    screen = true,      
    anonymous = false,   
]
{acmart}

\fi

\usepackage{amsxtra, amsfonts, amssymb, amstext}
\usepackage{amsthm}
\usepackage{booktabs}
\usepackage{longtable}
\usepackage{nicefrac}
\usepackage{xspace}
\usepackage{url}
\usepackage{graphics,color}
\usepackage[algo2e,ruled,vlined,linesnumbered]{algorithm2e}
\newcommand{\assign}{\leftarrow}
\usepackage{graphicx}
\usepackage{wrapfig}
\usepackage{balance}

\ifarxiv
\usepackage[noadjust]{cite}
\usepackage[colorlinks]{hyperref}
\definecolor{linkblue}{rgb}{0.1,0.1,0.8}
\hypersetup{colorlinks=true,linkcolor=linkblue,filecolor=linkblue,urlcolor=linkblue,citecolor=linkblue}
\usepackage{caption}
\usepackage{subcaption}

\fi

\newcommand{\ignore}[1]{}

\allowdisplaybreaks[1]

\newcommand{\R}{\mathbb{R}}

\renewcommand{\epsilon}{\varepsilon}

\newcommand{\onemax}{\textsc{OneMax}\xspace}

\newcommand{\OM}{\textsc{Om}\xspace}

\newcommand{\ga}{$(1 + (\lambda,\lambda))$~GA\xspace}

\DeclareMathOperator{\mut}{flip}
\DeclareMathOperator{\Bin}{Bin}
\DeclareMathOperator{\cross}{cross}
\DeclareMathOperator{\stat}{static}
\DeclareMathOperator{\dyn}{dyn}
\DeclareMathOperator{\nint}{nint}

\begin{document}
\title{Hyper-Parameter Tuning for the \texorpdfstring{$(1+(\lambda,\lambda))$~GA}{(1+(L,L))~GA}}

\ifgecco
\author{Nguyen Dang}
\affiliation{\institution{School of Computer Science, University of St-Andrews, UK}} 
\author{Carola Doerr}
\affiliation{\institution{LIP6, Sorbonne Universit\'e, CNRS, Paris, France}}
\renewcommand{\shortauthors}{N. Dang, C. Doerr}
\else
\author{Nguyen Dang$^1$ and Carola Doerr$^2$}
\date{
$^1$University of St Andrews, School of Computer Science, St Andrews, Scotland, UK\\
$^2$Sorbonne Universit\'e, CNRS, Laboratoire d'Informatique de Paris 6, Paris, France\\ 
\vspace{1.5ex}
\today
}
\fi

\ifarxiv
\maketitle
\fi

\begin{abstract}
It is known that the $(1+(\lambda,\lambda))$~Genetic Algorithm (GA) with self-adjusting parameter choices achieves a linear expected optimization time on OneMax if its hyper-parameters are suitably chosen. However, it is not very well understood how the hyper-parameter settings influences the overall performance of the $(1+(\lambda,\lambda))$~GA. Analyzing such multi-dimensional dependencies precisely is at the edge of what running time analysis can offer. To make a step forward on this question, we present an in-depth empirical study of the self-adjusting $(1+(\lambda,\lambda))$~GA and its hyper-parameters. We show, among many other results, that a 15\% reduction of the average running time is possible by a slightly different setup, which allows non-identical offspring population sizes of mutation and crossover phase, and more flexibility in the choice of mutation rate and crossover bias---a generalization which may be of independent interest. We also show indication that the parametrization of mutation rate and crossover bias derived by theoretical means for the static variant of the $(1+(\lambda,\lambda))$~GA extends to the non-static case.
\end{abstract}

\ifgecco
%
%


\copyrightyear{2019}
\acmYear{2019}
\setcopyright{acmlicensed}
\acmConference[GECCO '19]{Genetic and Evolutionary Computation Conference}{July 13--17, 2019}{Prague, Czech Republic}
\acmBooktitle{Genetic and Evolutionary Computation Conference (GECCO '19), July 13--17, 2019, Prague, Czech Republic}
\acmPrice{15.00}
\acmDOI{10.1145/3321707.3321725}
\acmISBN{978-1-4503-6111-8/19/07}

\maketitle
\fi

\sloppy{
\section{Introduction}%
\label{sec:introduction}

The $(1+(\lambda,\lambda))$~Genetic Algorithm (GA) is a crossover-based evolutionary algorithm that was introduced in~\cite{DoerrDE15} to demonstrate that the idea of recombining previously evaluated solutions can be beneficial also on very smooth functions. More precisely, it was proven in~\cite{DoerrDE15,DoerrD18ga} that the \ga achieves an $o(n \log n)$ expected optimization time on \onemax, the problem of maximizing functions of the type $f_z:\{0,1\}^n \to \R, x \mapsto |\{1 \le i \le n \mid x_i=z_i\}|$. All purely mutation-based algorithms, in contrast, are known to require $\Omega(n \log n)$ function evaluations, on average, to optimize these functions~\cite{LehreW12,DoerrDY16}. 

The \ga has three parameters, the population size $\lambda$ of mutation and crossover phase, the mutation rate $p$, and the crossover bias $c$. It was shown in~\cite{DoerrDE15} that an asymptotically optimal linear expected running time can be achieved by the \ga when choosing these parameters in an optimal way, which depends on the fitness of a current-best solution. This result was extended in~\cite{DoerrD18ga} to a self-adjusting variant of the \ga, which uses a fixed parametrization $p=\lambda/n$, $c=1/\lambda$, and an adaptive success-based choice of $\lambda$. More precisely, in the self-adjusting \ga the parameter $\lambda$ is chosen according to a one-fifth success rule, which decreases $\lambda$ to $\lambda/F$ if an iteration has produced a strictly better solution, and increases $\lambda$ to $F^{1/4}\lambda$ otherwise. This linear runtime result proven in~\cite{DoerrD18ga} was the first example where a self-adjusting choice of the parameter values could be rigorously shown to outperform any possible static setting. 

Despite these theoretically appealing results, the performances reported in the original work introducing this algorithm~\cite{DoerrDE15} are rather disappointing in that they are much worse than those of Randomized Local Search for all tested problem dimensions up to $n=5\,000$. It was pointed out in~\cite{CarvalhoD18} that this is partially due to a sub-optimal implementation; the average optimization times reduce drastically when enforcing that at least one bit is flipped in the mutation phase. In this case, the self-adjusting \ga starts to outperform RLS already for dimensions around $1\,000$. Another possible reason 
lies in the fact that the hyper-parameters of the self-adjusting \ga had not been optimized. In~\cite{DoerrDE15} the authors had taken some default values from the literature, and show only some very basic sensitivity analysis with respect to the update strength, but not with respect to any of the other parameters such as the success rate. In~\cite{DoerrD18ga} some general advice on choosing the hyper-parameters is given, but their influence on the explicit running time is not discussed, mostly due to missing precision in the available results, which state the asymptotic linear order only, but not the leading constants or lower order terms. Also the update strength $F$ for which the linear running time is obtained is only shown to exist, but not made explicit in~\cite{DoerrD18ga}. 

To shed light on the question how much performance can be gained by choosing the hyper-parameters of the \ga with more care, we present in this work a detailed empirical evaluation of this parameter tuning question. Our first finding is that the default setting studied in~\cite{DoerrDE15}, which uses update strength $F=3/2$ and the mentioned $1/5$-th success rule is almost optimal. More precisely, we show that for all tested problem dimensions between $n=500$ and $n=10\,000$ only marginal gains are possible by choosing different update strengths $F$ and/or a success rule different from 1/5. 

We then introduce a more general variant of the \ga, in which the offspring population sizes of mutation and crossover phase need not be identical, and in which more flexible choices of mutation strength and crossover bias are possible. This leaves us with a five-dimensional hyper-parameter tuning problem, which we address with the irace software~\cite{irace}. We thereby find configurations whose average optimization times are around $15\%$ better than that of the default self-adjusting \ga, for each of the tested dimensions. The configurations achieving these advantages are quite stable across all dimensions, so that we are able to derive configurations achieving these gains for all dimensions. We furthermore show that the relative advantage also extends to dimensions $20\,000$ and $30\,000$, for which we did not perform any hyper-parameter tuning. 
This five-dimensional variant of the \ga is also of independent interest, since it allows much greater flexibility than the standard versions introduced in~\cite{DoerrDE15,DoerrD18ga}.

We finally study if hyper-parameter tuning of a similarly extended static \ga can give similar results, or whether the asymptotic discrepancy between non-static and static parameter settings proven in~\cite{DoerrD18ga} also applies relatively small dimensions. We show that indeed already for the smallest tested dimension, $n=500$, the average optimization time of the best static setting identified by our methods is around 5\% worse than the standard self-adjusting \ga from~\cite{DoerrDE15,DoerrD18ga}, and by 22\% worse than the best found five-dimensional configuration. This disadvantage increases to $22\%$ and $45\%$ in dimension $n=10\,000$, respectively, thus showing that not only the advantage of the self-adjusting \ga kicks in already for small dimensions, but also confirming that the relative advantage increases with increasing problem dimensions. 

Apart from introducing the new \ga variants, which offer much greater flexibility than the standard versions, 
our work significantly enhance our understanding of the hyper-parameter setting in the \ga, paving the way for a precise rigorous theoretical analysis. In particular the stable performance of the tuned configurations indicates that a precise running time analysis might be possible. We furthermore learn from our work that the parametrization of the mutation rate and the crossover bias, which were suggested and proven to be asymptotically optimal for the static case in~\cite{DoerrD18ga}, seem to be optimal also in the non-static case with self-adjusting parameter choices. Finally, we also observe that for the generalized dynamic setting $1:x$ success rules with success rates between $3$ to $4$ seem to be slightly better than the classic one-fifth success rule with $F=3/2$. 

\textbf{Broader Context: Parameter Control and Hyper-Parameter Tuning.} All iterative optimization heuristics such as EAs, GAs, local search variants, etc. are parametrized algorithms. Choosing the right parameter values is a tedious, but important task, frequently coined the ``Achilles' heel of evolutionary computation''~\cite{FialhoCSS10}. It is well known that choosing the parameter values of different parameter settings can result in much different performances. Extreme cases in which a small constant change in the mutation rate result in super-polynomial performance gaps were shown, for example, in~\cite{DoerrJSWZ13,Lengler18}. 

To guide the user in the parameter selection task, two main approaches have been developed: parameter tuning and parameter control. \emph{Parameter tuning} aims at developing tools that automatize the process of identifying reasonable parameter values, cf.~\cite{SMAC,irace,Hyperband,ParamILS,GGA} for examples. \emph{Parameter control}, in contrast, aims to not only identify such good values, but to also track the evolution of good configurations during the whole optimization process, thereby achieving additional performance gains over an optimally tuned static configuration, cf.~\cite{KarafotiasHE15,AletiM16,DoerrD18chapter} for surveys. In practice, parameter control mechanisms are parametrized themselves, thus introducing \emph{hyper-parameters,} which again need to be chosen by the user or one of the tuning tools mentioned above. This is also the route taken in this present work: in Sections~\ref{sec:dyn01} and~\ref{sec:dynamic} we will use the iterated racing algorithm \emph{irace}~\cite{irace} to tune two different sets of hyper-parameters of the self-adjusting \ga, a two-dimensional and a five-dimensional one. In Section~\ref{sec:static} we then tune the four parameters of a generalized static \ga variant. By comparing the results of these tuning steps, we obtain the mentioned estimates for the relative advantage of the self-adjusting over the best tuned static parameter configuration. 

\textbf{Reproducibility, Raw Data, and Computational Resources.} 
\ifgecco
Given the space limitations, we only display selected statistics.
\fi
We concentrate on reporting average values to match with the available theoretical and empirical results. We recall that in theoretical works the expected optimization time dominates all other performance measures. Selected boxplots for the most relevant configurations are provided in Section~\ref{sec:distribution}. 
\ifarxiv
Information about the selected parameter values and the empirical quantiles of the running times can be found in the appendix. 
\fi
Source codes, additional performance statistics, summarizing plots, heatmaps with different colormaps, and raw data can be found on our GitHub repository at~\cite{Supplementary}. All experiments were run on the HPCaVe cluster~\cite{HPCaVe}, whose each node consists of two 12-core Intel Xeon E5 2.3GHz with 128Gb memory.

\section{Tuning the default \texorpdfstring{\ga}{(1+(l,l)) GA}}%
\label{sec:dyn01}

Our main interest is in tuning the self-adjusting variant of the \ga proposed in~\cite{DoerrDE15} and analyzed in~\cite{DoerrD18ga}. As in these works, we regard the performance of this algorithm on the \onemax problem. The \onemax problem is one of the most classic benchmark problems in the evolutionary computation literature. It asks to find a secret string $z$ via calls to the function $f_z:\{0,1\}^n \to \R, x \mapsto |\{1 \le i \le n \mid x_i=z_i\}|$ and is thus identical to the problem of minimizing the Hamming distance to an unknown string $z \in \{0,1\}^n$. It is referred to as ``\onemax'' in evolutionary computation, since the performance of most EAs (including the \ga) is identical on any of the functions $f_z$, and it therefore suffices to study the instance $f_{(1,\ldots,1)}$.  

It is known that the best possible mutation-based (i.e., formally, the best unary unbiased) black-box algorithms have an expected optimization time on \onemax of order $n \log n$~\cite{LehreW12,DoerrDY16}. The \ga, in contrast, achieves a linear expected optimization time if its parameters are suitably chosen~\cite{DoerrDE15,DoerrD18ga}. Parameter control, i.e., a non-static choice of these parameters, is essential for the linear performance, since the \ga with static parameter values cannot have an expected optimization time that is of better order than $n \sqrt{\log(n) \log\log\log(n) / \log\log(n)}$, which is super-linear.

\ifgecco
\setlength{\intextsep}{1\baselineskip}
\fi
\begin{algorithm2e}[!h]%
\textbf{Initialization:} 
		Sample $x \in \{0,1\}^n$ u.a.r.\;
		Initialize $\lambda \assign 1$\;
\textbf{Optimization:}
\For{$t=1,2,3,\ldots$}{
\underline{\textbf{Mutation phase:}}\\
\Indp
	Sample $\ell$ from $\Bin_{>0}(n,p=\alpha \lambda/n)$\;
	\lFor{$i=1, \ldots, \lambda_1=\nint(\lambda)$}{$x^{(i)} \assign \mut_{\ell}(x)$}
	Choose $x' \in \{x^{(1)}, \ldots, x^{(\lambda_1)}\}$ with $f(x')=\max\{f(x^{(1)}), \ldots, f(x^{(\lambda_1)})\}$ u.a.r.\;
\Indm
\underline{\textbf{Crossover phase:}}\\
\Indp
\lFor{$i=1, \ldots, \lambda_2=\nint(\beta \lambda)$}{$y^{(i)} \assign \cross_{c=\gamma/\lambda}(x,x')$}
Choose $y \in \{x',y^{(1)}, \ldots, y^{(\lambda_2)}\}$ with $f(y)=\max\{f(x'),f(y^{(1)}), \ldots, f(y^{(\lambda_2)})\}$ u.a.r.\;
\Indm
\underline{\textbf{Selection and update step:}}\\
\Indp
\lIf{$f(y)>f(x)$}{
$x \assign y$; $\lambda \assign \max\{b\lambda,1\}$}
\lIf{$f(y)=f(x)$}{
$x \assign y$; $\lambda \assign \min\{A\lambda,n-1\}$}
\lIf{$f(y)<f(x)$}{$\lambda \assign \min\{A\lambda,n-1\}$}
\Indm
}
\caption{The self-adjusting \ga variant $\dyn(\alpha,\beta,\gamma,A,b)$ with five hyper-parameters.}
\label{alg:dyn}
\end{algorithm2e}
\ifgecco
\setlength{\intextsep}{1\baselineskip}
\fi

\subsection{The dynamic \texorpdfstring{\ga $\dyn(\alpha,\beta,\gamma,A,b)$}{(1+(l,l)) GA}}
The \ga is a binary unbiased algorithm, i.e., it applies crossover but uses only variation operators that are invariant with respect to the problem representation. We present the pseudo-code of the \ga in Algorithm~\ref{alg:dyn}, in which we denote by $\nint(.)$ the nearest integer function, i.e., $\nint(r)=\lfloor(r)\rfloor$ if $r-\lfloor r \rfloor <1/2$ and $\nint(r)=\lceil(r)\rceil$ otherwise.

The \ga has two phases, a mutation phase and a crossover phase, followed by a selection step. 
In the \emph{mutation phase} $\lambda_1=\nint(\lambda)$ offspring are evaluated. Each of them is sampled by the operator $\mut_\ell(.)$ uniformly at random (u.a.r.) from all the points at a radius $\ell$ around the current-best solution $x$. The radius $\ell$ is sampled from the conditional binomial distribution $\Bin_{>0}(n,p)$, which assigns to each positive integer $1 \le k \le n$ the probability $\Bin_{>0}(n,p)=\binom{n}{k}p^k(1-p)^{n-k}/(1-(1-p)^n)$. Following the reasoning made in~\cite{CarvalhoD18} we deviate here from the \ga variants investigated in~\cite{DoerrDE15}, to avoid useless iterations. The variants  analyzed in~\cite{DoerrDE15,DoerrD18ga} allow $\ell=0$, which is easily seen to create copies of the parent only. As it cannot advance the search, we enforce $\ell \ge 1$.  

In the \emph{crossover} phase, $\lambda_2$ offspring are evaluated. They are sampled by the crossover operator $\cross_c(\cdot,\cdot)$, which creates an offspring by copying with probability $c$, independently for each position, the entry of the second argument, and by copying from the first argument otherwise. We refer to the parameter $0<c<1$ as the \emph{crossover bias}. Again following~\cite{CarvalhoD18}, we evaluate only those offspring that differ from both their two parents; i.e., offspring that are merely copies of $x$ or $x'$ do not count towards the cost of the algorithm, since their function values are already known. 

In the \emph{selection step,} we replace the parent by its best offspring if the latter is at least as good. When a strict improvement has been found, the value of $\lambda$ is updated to $\max\{b \lambda,1\}$. It is increased to $\min\{A \lambda,n-1\}$ otherwise.
 
Note that in the description above and Algorithm~\ref{alg:dyn} we have deviated from the commonly used representation of the \ga, in that we have parametrized the mutation rate as $p=\alpha \lambda/n$, the offspring population size of the crossover phase as $\lambda_2=\nint(\beta \lambda)$, the crossover bias as $c=\gamma/\lambda$, and in that we allow more flexible update strengths $A$ and $b$. We thereby obtain a more general variant of the \ga, which we will show to outperform the standard self-adjusting one considerably.  In this present section, however, we only generalize the update rule, not yet the other parameters. That is, we work in this section only with the \ga variant $\dyn(1,1,1,A,b)$, which uses $\lambda_1=\lambda_2$, $p=\lambda/n$, and $c=1/\lambda$.

In our implementation we always ensure that $p$ and $c$ are at least $1/n$ and at most $0.99$, by capping these values if needed. Slightly better performances may be obtained by allowing even smaller $p$-values, but we put this question aside for this present work. 

\ifgecco
\begin{figure}
\centering
\includegraphics[width=0.9\linewidth]{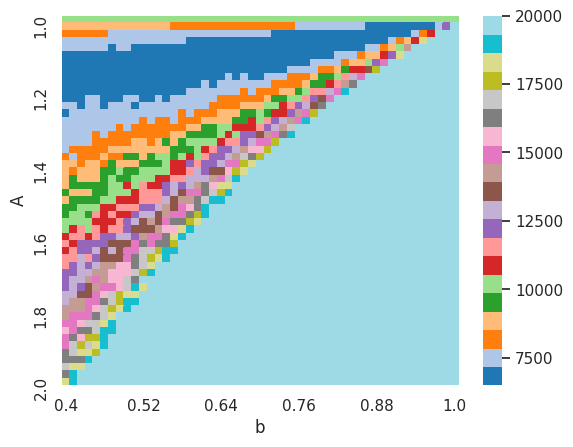}
\caption{Heatmap for $\dyn(\alpha=\beta=\gamma=1, A\in [1.02,2], b\in [0.4,0.988])$, average optimization time capped at 20\,000}
\label{fig:heatmap2}
\Description{The description of this plot is given in the main text.}
\end{figure}
\else
\begin{figure*}
\centering
\begin{subfigure}[t]{0.46\textwidth}
	\centering
	\includegraphics[width=\textwidth]{heatmap-1000-boundMax-20000-mean}
\end{subfigure}%
    ~ 
    \begin{subfigure}[t]{0.46\textwidth}
    \centering
		\includegraphics[width=\textwidth]{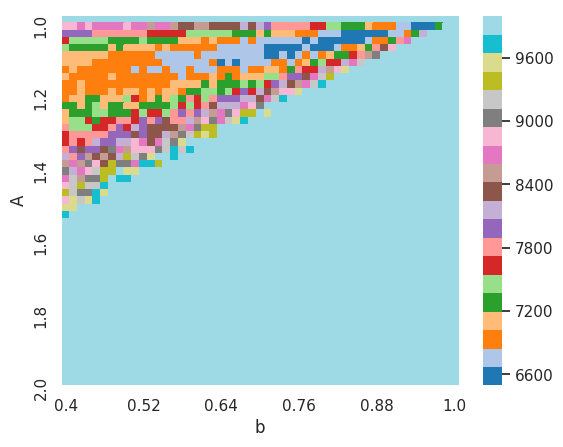}
\end{subfigure}
\caption{Heatmap with average optimization times of $\dyn(\alpha=\beta=\gamma=1,A,b)$ with $A\in [1.02,2], b\in [0.4,0.988]$ and capping at 20\,000 evaluations in Fig.~(a) and zooming into $A\in [1.02,1.6], b\in [0.4,0.988]$ and budget capped at 10\,000 evaluations in Fig.~(b)}
\label{fig:heatmap2}
\label{fig:heatmap}
\end{figure*}
\fi

\subsection{Influence of the Update Strengths}
\label{sec:heatmap}

As mentioned above, in our first set of experiments we focus on investigating the influence of the update strengths $A$ and $b$, i.e., we fix $\alpha=\beta=\gamma=1$ in the notation of Algorithm~\ref{alg:dyn}. In~\cite{DoerrDE15} it was suggested to set $A=(3/2)^{1/4} \approx 1.11$ and $b=2/3$. These settings had previously been suggested in~\cite{Auger09,KernMHBOK04} in a much different context, but seemed to work well enough for the purposes of~\cite{DoerrDE15} and was hence not questioned further in that work (apart from a simple evaluation showing that for $n=400$ the influence of varying the update strength $F$ within the interval $[1.1,2]$ is not very pronounced). Note that the choices of $A$ and $b$ correspond to an implicit one-fifth success rule, in the sense that the value of $\lambda$ is stable if one out of five iterations is successful. The \emph{success rate} (five in this case) can be computed as $1-\ln(b)/\ln(A)$. We emphasize that for notational convenience we prefer to speak of a success rate $x$ instead of a $1/x$-th success rule.

The heatmap in Figure~\ref{fig:heatmap2} shows the average running time of the self-adjusting \ga in dependence of the update strengths $A$ and $b$. We considered all combinations of 50 equally spaced values for $A \in \{1.02, 1.04, \ldots, 2\}$ and for $b \in \{0.4, 0.412, \ldots, 0.988\}$ (2\,500 hyper-parameter settings). For each setting, we performed 100 independent runs of the algorithm $\dyn(1,1,1,A,b)$. Each run has a maximum budget of 150\,000 function evaluations. Our results are for problem dimension $n=1\,000$. 
\ifgecco
To show more structure, we cap in Figure~\ref{fig:heatmap2} the values at $20\,000$, other versions with different color schemes and cappings are available at~\cite{Supplementary}. 
\else
To show more structure, we cap in Figure~\ref{fig:heatmap2}.~(a) the values at $20\,000$. A zoom into the interesting region of combinations achieving an average optimization time $\le 10\,000$ can be found in Figure~\ref{fig:heatmap}.(b). More versions with different color schemes and cappings are available at~\cite{Supplementary}.
\fi

The best configuration is $(A=1.06, b=0.82)$ with an estimated average optimization time of $6\,495$. This configuration has a success rate of $4.4$. The average optimization time of the default variant $\dyn(1,1,1,(3/2)^{1/4},2/3)$ from~\cite{DoerrDE15}, denoted by $\dyn(\text{default})$ in the following, over 500 runs is $6,671$, and thus only $2.7\%$ worse than $\dyn(1,1,1,1.06,0.82)$. $29$ of the $2\,500$ tested configurations have a smaller average optimization time than $\dyn(\text{default})$, all of them with $A$-values at most $1.12$ and $b$-value at least $0.64$. 106 configurations are worse by at most 3\%, and 188 by at most 5\%.

For a more stable comparison, we also ran $\dyn(1,1,1,1.06,0.82)$ 500 times, and its average optimization time increased to $6\,573$ for these 500 independent runs, reducing the relative advantage over $\dyn(\text{default})$ to $1.5\%$. Boxplots with information about the runtime distributions can be found in Section~\ref{sec:distribution}.

\begin{figure}
\centering
\includegraphics[width=0.9\linewidth]{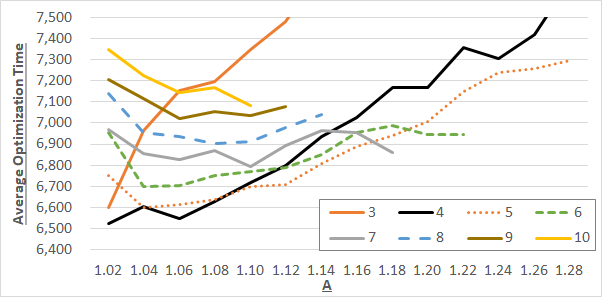}
\caption{Average optimization time for different success rates, sorted by value of $A$}
\label{fig:success-rate}
\ifgecco
\Description{The description of this plot is given in the main text.}
\fi
\end{figure}

In Figure~\ref{fig:success-rate} we plot the average optimization time for different success rates, sorted by the value $A$. Note that for each tested $A$-value we have averaged here over all configurations using the same rounded (by $\nint(\cdot)$) success rate. The performance of success rates 1 and 2 is much worse than $7\,500$ and is therefore not plotted. We plot only results for success rates at most 10, for readability purposes. We see that success rates 4 and 5 are particularly efficient, given the proper values of $A$. The performance curves for success rates $\ge 4$ seem to be roughly U-shaped with different values of $A$ in which the minimum is obtained. It could be worthwhile to extend the mathematical analysis of the $\dyn(1,1,1,A,b)$ presented in~\cite{DoerrD18ga} in order to identify the precise relationship.

\subsection{Tuning with irace}
\label{sec:irace}

The computation of the heatmaps presented above is quite resource-consuming, around 286 CPU days for the full heatmap with $2\,500$ parameter combinations for $n=1\,000$. Since we are interested in studying the quality of the \ga also for other problem dimensions, we therefore investigate how well automated tuning tools approximate the best known configuration. To this end, we run the configuration tool irace ~\cite{irace} with \emph{adaptive capping}~\cite{iracecapping} enabled. This new mechanism was recently added to irace to make its search procedure more efficient when optimizing running time or time-compatible performance measurement. We use irace to optimize the configuration of the $\dyn(1,1,1,A,b)$ for values of $A$ between $1$ and $2.5$, and values of $b$ between $0.4$ and $1$. The allocated budget is 10 and 20 hours of walltime on one 24-core cluster node for $n \le 5\,000$ and $n > 5\,000$, respectively. This time budget is only a fraction of the ones required by heatmaps (around $3\%$ for $n=1\,000$).

\begin{figure}[t]
\centering
\includegraphics[width=0.9\linewidth]{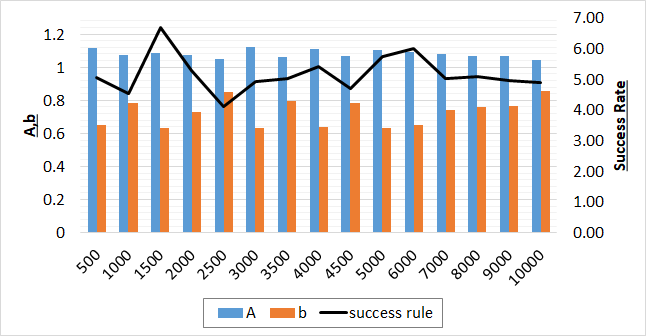}
\caption{Parameter values suggested by irace for the \ga variant $\dyn(1,1,1,A,b)$. The success rate equals $1-\ln(b)/\ln(A)$}
\label{fig:paramAb}
\ifgecco
\Description{The description of this plot is given in the main text.}
\fi
\end{figure}

For $n=1\,000$ irace suggests to use configuration $(A=1.071, b=0.7854)$, which is similar to the one showing best performance in the heatmap. The average optimization time of this configuration is $6,573$ (this number, like all numbers for the configurations suggested by irace are simulated from 500 independent runs each), and thus identical to the best one from the heatmap computations. The suggested configuration corresponds to a $4.52$ success rate. 

\begin{figure*}
\centering
\ifgecco
\includegraphics[width=0.9\linewidth]{runtime-summary-GECCO}
\else
\includegraphics[width=\linewidth]{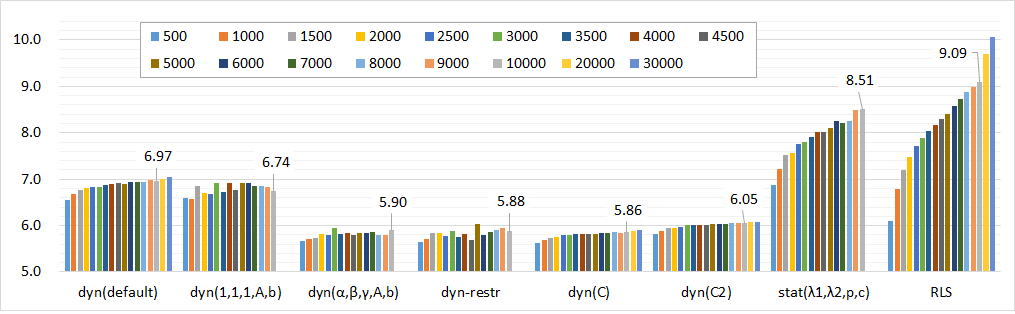}
\fi
\caption{By $n$ normalized average optimization times for 500 independent runs each. For data sets $\dyn(1,1,1,A,b)$, $\dyn(\alpha,\beta,\gamma,A,b)$, 
and $\stat(\lambda_1,\lambda_2,p,c)$ we have taken for each dimension the configuration suggested by irace; the other results are for fixed configurations. Displayed numbers are for $n=10\,000$.}
\label{fig:runtime-summary}
\ifgecco
\Description{The description of this plot is given in the main text.}
\fi
\end{figure*}

Confident that irace is capable of identifying good parameter settings, we then run irace for various problem dimensions between $500$ and $10\,000$. The by $n$ normalized average optimization time of the suggested configurations are reported in Figure~\ref{fig:runtime-summary} in column $\dyn(1,1,1,A,b)$. The chosen $A$-values are between $1.04$ and $1.12$ and the $b$-values are between $0.63$ and $0.88$, with corresponding success rates between $4.41$ and $6.68$, cf. Figure~\ref{fig:paramAb}. We observe a quite stable suggestion for the parameter values. 
\ifarxiv
The suggested configurations, along with statistical data can be found in Table~\ref{tab:dyn01-stats} in the appendix.
\else
\fi

In Figure~\ref{fig:runtime-summary} we also display, in column $\dyn(\text{default})$, the normalized average optimization times of the default setting $(1,1,1,(3/2)^{1/4},2/3)$. The relative disadvantage of the $\dyn(1,1,1,A,b)$ over the $\dyn(\text{default})$ ranges from $-1.3\%$ to $3.3\%$. The negative values (in four dimensions) may be due to a suboptimal suggestion of irace, or due to the variance of the algorithms; the relative standard deviation is between $5\%$ and $10\%$, cf. also the boxplots in Section~\ref{sec:boxplots}. 

We also observe that the normalized average optimization times of $\dyn(\text{default})$ increase slightly with increasing problem dimension. Note, however, that this does not necessarily tell us something about the constant factor in the linear running time of this algorithm, although the results indicate that this factor might be larger than $7$. Already for $n=1\,000$ the $\dyn(\text{default})$ has a smaller average optimization time than RLS, the relative advantage of $\dyn(\text{default})$ is around $2\%$, and increases to around $31\%$ for $n=30\,000$.

\section{5-dimensional Parameter Tuning}%
\label{sec:dynamic}

Next we turn our attention to the five-dimensional \ga variant $\dyn(\alpha,\beta,\gamma,A,b)$, in which not only the update strengths $A$ and $b$ are configurable, but also the dependence of $p=\alpha \lambda /n$, $\lambda_2=\nint(\beta \lambda)$, $c=\gamma/\lambda$. The dependencies of the parameters on $\lambda$ are based on a theoretical result proven in~\cite{DoerrD18ga}, where it is shown that any static configuration with $\lambda_2=\lambda_1$ (i.e., $A=b=\beta=1$) that achieves optimal asymptotic expected performance must necessarily satisfy $p=\Theta(\lambda/n)$ and $\gamma=\Theta(1/\lambda)$. 

To investigate how much performance can be gained by this flexibility, and how reasonable parameter values look like, we run again irace, this time using the following parameter ranges: $\alpha \in (1/3,10)$, $\beta \in (1,10)$, $\gamma \in (1/3,10)$, $A \in (1.01,2.5)$ and $b \in (0.4,0.99)$. The allocated budget is the same as for the $\dyn(1,1,1,A,b)$, i.e., 240 CPU hours for $n \le 5\,000$ and 480 CPU hours for $n > 5\,000$.

The normalized average running times of the suggested configurations are presented in Column $\dyn(\alpha,\beta,\gamma,A,b)$ in Figure~\ref{fig:runtime-summary}. We observe that the parametrization of $\lambda_2$, $p$, and $c$ consistently allows to decrease the average optimization time by around 14\%, when measured against the best $\dyn(1,1,1,A,b)$ variant. 

\subsection{Suggested Hyper-Parameters}

\begin{figure}
\centering
\includegraphics[width=\linewidth]{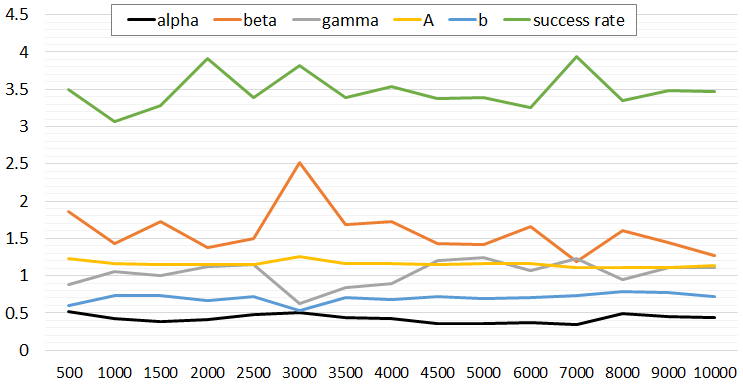}
\caption{Hyper-parameters and success rate suggested by irace for the  $\dyn(\alpha,\beta,\gamma,A,b)$ configuration problem.}
\label{fig:dyn02-param02-parameter-by-n}
\ifgecco
\Description{The description of this plot is given in the main text.}
\fi
\end{figure}

The suggested parameter values are displayed in Figure~\ref{fig:dyn02-param02-parameter-by-n}. We observe that these are quite stable, in particular when ignoring the $3\,000$ and $7\,000$ dimensional configurations. More precisely, irace consistently suggests configurations with $\alpha \approx 0.45$, $\beta \approx 1.6$, $\gamma \approx 1$, $A \approx 1.16$, and $b \approx 0.7$, with corresponding success rates between $3$ and $4$. These stable values suggest that the parametrization chosen in Algorithm~\ref{alg:dyn} (and originally derived in~\cite{DoerrD18ga} for the static \ga) is indeed suitable also for the non-static setting.

\begin{figure*}
\centering
\includegraphics[width=0.9\linewidth]{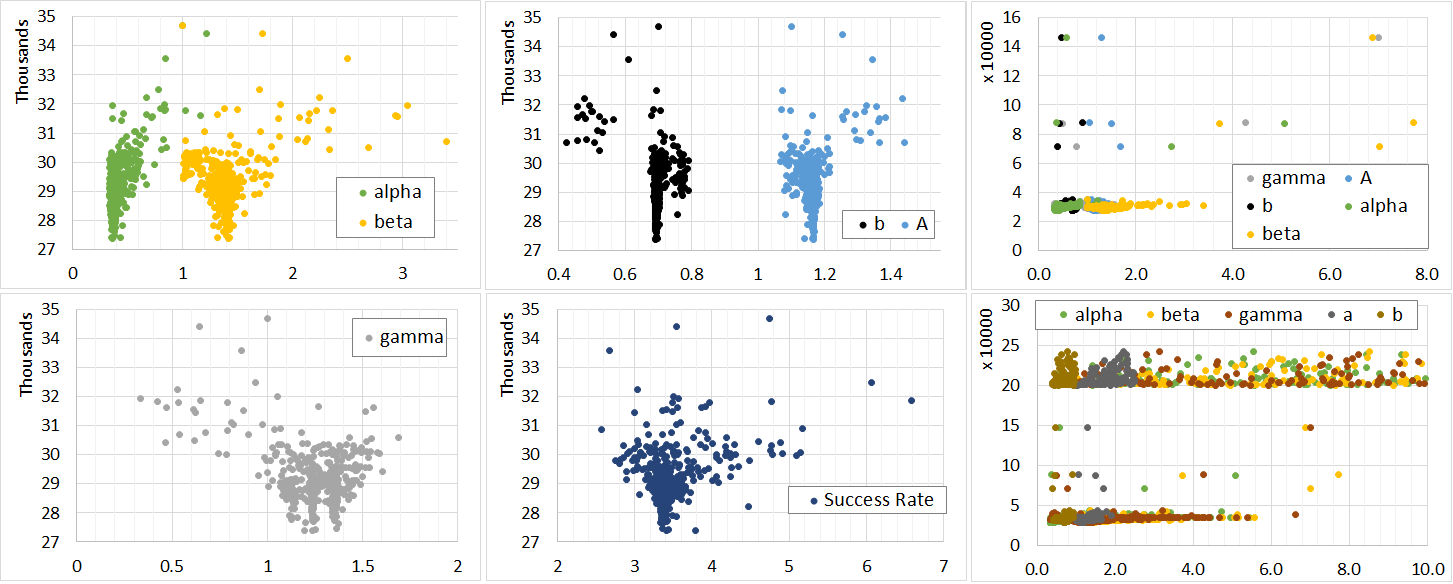}
\caption{Average running times of different configurations on $5\,000$-dimensional \onemax. See text for a description.}
\label{fig:dynam02param02-RT-per-param}
\ifgecco
\Description{The description of this plot is given in the main text.}
\fi
\end{figure*}

In Figure~\ref{fig:dynam02param02-RT-per-param} we plot the average optimization time of the configurations tested by irace for $n=5\,000$ in dependence of each of the five hyper-parameters $\alpha,\beta,\gamma,A,b$ and in dependence of the success rate $1-\ln(b)/\ln(A)$. Note that the number of runs differs from point to point, depending on how many evaluations irace has performed for each of these configurations. It is important to note that the capping procedure may stop an algorithm before it has found an optimal solution, in order to save time for the evaluation of more promising configurations. 
The plotted values are the averages of the successful runs only. An exception to this rule is the chart on the lower right, which shows the whole range of all $2\,212$ tested configurations; these values are the average time after which the configurations had either found the optimum or were stopped by the capping procedure. We thus see that irace has indeed tested across the whole range of admitted parameter values. Around $38\%$ of all $4\,961$ runs were stopped before an optimum had been found. However, we already see here that for each parameter there are configurations which use a good value for this parameter, but which shows quite poor overall performance. These results indicate that no parameter alone explains the performance, but that interaction between different parameter values is indeed highly relevant; we will discuss this aspect in more detail below. 

Out of the $2\,212$ tested configurations only $765$ configurations had at least one successful run. The averages of all successful runs are plotted in the upper right chart of Figure~\ref{fig:dynam02param02-RT-per-param}. We observe that the well-performing region of values for each parameter is quite concentrated. The charts on the left and in the middle column zoom into those configurations which had an average optimization time smaller than $35\,000$. These plots give a good indication where the interesting regions for each parameter are. We also plot the average optimization time in dependence of the success rate and see good performance for success rates between 3 and 4. 

For $348$ tested configurations only successful runs were reported; i.e., for these configurations none of the runs had been stopped before it had found an optimal solution. 
 When restricting the zoomed plots in Figure~\ref{fig:dynam02param02-RT-per-param} to only those $348$ configurations, we obtain a very similar picture. 
We omit a detailed discussion but note that these plots can be found in our repository~\cite{Supplementary}. 

The final configuration suggested by irace, $\dyn(0.3594, 1.4128, 1.2379, 1.1672, 0.691)$ has an average optimization time of $29\,165$ in the 500 independent runs conducted for the values reported in Figure~\ref{fig:runtime-summary}. During the irace optimization the estimated average was $28\,876$ (across $50$ runs). 

We see that some of the configurations in Figure~\ref{fig:dynam02param02-RT-per-param} have a smaller average optimization time than this latter value. In fact, there are 292 such configurations with at least one successful run and $62$ configurations with only successful runs. As we can see from the plots in Figure~\ref{fig:dynam02param02-RT-per-param} all these configurations have very similar parameter values. This observation nevertheless raises the question why irace has not suggested one of these presumably better configurations instead. To understand this behavior, we investigate in more detail the working principles of irace, and find two main reasons. One is that the time budget did not allow a further investigation of these configurations, so that statistical evidence that they are indeed superior to the suggested one was not sufficient. A second reason is that the capping suggested in~\cite{iracecapping} resulted in a somewhat harsh selection of ``surviving'' configurations. We leave the question if any of the 292 configurations would have been significantly better than the suggested one for future work. Overall, our investigation suggests that some adjustments to irace's default setting might be useful for applications similar to ours, where the performance measure may potentially suffer from high variance.

We next investigated the influence of each parameter on the overall running time. To this end, we have applied \emph{the functional analysis of variance (fANOVA)}~\cite{hoos2014efficient} on the performance data given by irace. fANOVA can efficiently recognize the importance of both individual algorithm parameters and their interactions through their percentage of contributions on the total performance variance. The software PyImp~\cite{PyImp} is used for the analysis. Obtained results are quite consistent among different dimensions. The most important parameter is $\alpha$, which explains on average $57\%$ of the total variance. The second most important parameter is $\gamma$, explaining around $22\%$ of the total variance, on average. Other important effects include pairwise interaction between $\alpha$ and $\gamma$ or $A$. Individual parameters and their pairwise interaction effects are able to explain almost $100\%$ of the total variance, so that there is no need to consider higher-order interactions. 

\ifarxiv
In light of the quite stable parameter values suggested by irace (Figure~\ref{fig:dyn02-param02-parameter-by-n}) one might hope to obtain even better results when restricting the ranges of possible parameter values further. To investigate this question we run irace again on the $\dyn(\alpha,\beta,\gamma,A,b)$ configuration problem, this time with restricted parameter ranges $\alpha \in (1/3,4)$, $\beta \in (1,3)$, $\gamma \in (1/3,4)$, $A \in (1.01,2)$ and $b \in (0.4,0.99)$. The normalized average running time results are reported in column dyn-restr. of Figure~\ref{fig:runtime-summary}. We observe that the advantage is negligible, and in four of the tested dimensions the suggested configurations even have a slightly worse average optimization time. This effect is likely to be caused by the randomness of the running times and/or the irace procedure itself.   
\fi

Finally, we derive from the suggested parameter values two configurations that we investigate in more detail: $\dyn(0.45,1.6,1,1.16,0.7)$ and $\dyn(1/2,2,1/2,(3/2)^{1/4},2/3)$, which we abbreviate as dyn(C) and dyn(C2), respectively. 
While dyn(C) consistently shows better performance than dyn(C2), the latter might be easier to analyze by theoretical means. Their normalized average optimization time across all tested dimensions can be found again in Figure~\ref{fig:runtime-summary}. They are considerably better than that of $\dyn(\text{default})=\dyn(1,1,1,(3/2)^{1/4},2/3)$, between $14\%$ and $16\%$ across all tested dimensions for dyn(C) and between $11\%$ and $13\%$ for dyn(C2). dyn(C2) is between $1\%$ and $4\%$ worse than the (for each dimension independently tuned) best suggested $\dyn(\alpha,\beta,\gamma,A,b)$ configuration. For dyn(C) we even observe that the average running times for the 500 runs are smaller than those of $\dyn(\alpha,\beta,\gamma,A,b)$ for 10 out of the 15 tested dimensions. The advantages of dyn(C) and dyn(C2) over $\dyn(\text{default})$ also translate to larger dimensions, for which we did not perform hyper-parameter tuning. 
For $n=20,000$ and $n=30,000$ the advantage of dyn(C) over $\dyn(\text{default})$ are 16\% each, and for dyn(C2) a relative advantage of 14\% is observed. 

\subsection{Fixed-Target Analysis}
\label{sec:FT}

Finally, we address the question where the advantage of the self-adjusting \ga over RLS originates from. To this end we perform an empirical fixed-target runtime analysis for two selected configurations, the default configuration $\dyn(\text{default})$ and the configuration dyn(C) mentioned above. 

The fixed-target running times have been computed with IOHprofiler~\cite{IOHprofiler}, a recently announced tool which automates the performance analysis of iterative optimization heuristics. 
The average results of $100$ independent runs for $n=10\,000$ are shown in Figure~\ref{fig:FT}. We observe that RLS is significantly better for almost all target values. In fact, the configuration dyn(C) has better first hitting times than RLS only for \onemax values greater than $9\,978$, i.e., only for the last 22 target values. We recall from Figure~\ref{fig:runtime-summary} that the average optimization time of dyn(C) is better than that of RLS by around $36\%$ for $n=10\,000$. To study at which point dyn(C) starts to perform better than RLS, we compute the gradient of the curves plotted in Figure~\ref{fig:FT}, showing that this happens around target value $9\,750$. For the default configuration $\dyn(\text{default})$ the situation is as follows: It is has smaller first hitting time than RLS only for target values $\ge 9\,995$, although its overall average running time is smaller by around 23\%. The gradient of $\dyn(\text{default})$ is better than that of RLS starting at target value around $9\,850$. Finally, dyn(C) has smaller average hitting time than $\dyn(\text{default})$ for $\OM$-values at least $8,934$, and a better gradient starting at around $8\,370$. We show in Figure~\ref{fig:FT} the hypothetical running times of an algorithm that runs RLS until target value $\OM(x)=9\,750$ and then switches to dyn(C). Its average running time is $17\%$ smaller than that of dyn(C), raising the interesting question how to detect such switching points on the fly. 

\begin{figure}
\centering
\includegraphics[width=\linewidth]{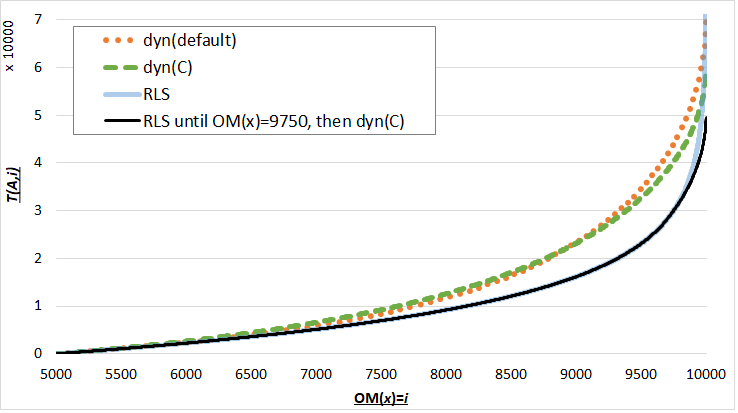}
\caption{Average fixed-target running times for RLS and two selected $\dyn(\alpha,\beta,\gamma,A,b)$ configurations, capped at 7,100 function evaluations. 
}
\label{fig:FT}
\ifgecco
\Description{The description of this plot is given in the main text.}
\fi
\end{figure}

\ifarxiv
We also want to understand how the value of $\lambda$ evolves during the optimization process. We plot in Figure~\ref{fig:param-lambda} for each target value the average value of $\lambda_1$ in the iteration in which for the first a solution of this quality has been sampled. More precisely, we display in Figure~\ref{fig:param-lambda} the logarithm of these parameter values. These results have again been computed with IOHprofiler~\cite{IOHprofiler}. We also show in this plot the logarithm of $\lambda^*=\sqrt{n/(n-\OM(x))}$, the value for which the standard dynamic \ga (with $\alpha=\beta=\gamma=1$) was first shown to have linear optimization time, cf.~\cite{DoerrDE15}. We observe that for both configurations the average value of $\lambda$ increases with already obtained function value, with a final value of $95$ for $\dyn(\text{default})$ and $117$ for dyn(C). We recall that the number of offspring generated in each iteration is $2\lambda_1$ for $\dyn(\text{default})$ and around $2.6 \nint(\lambda)$ for dyn(C). 

\begin{figure}
\centering
\includegraphics[width=\linewidth]{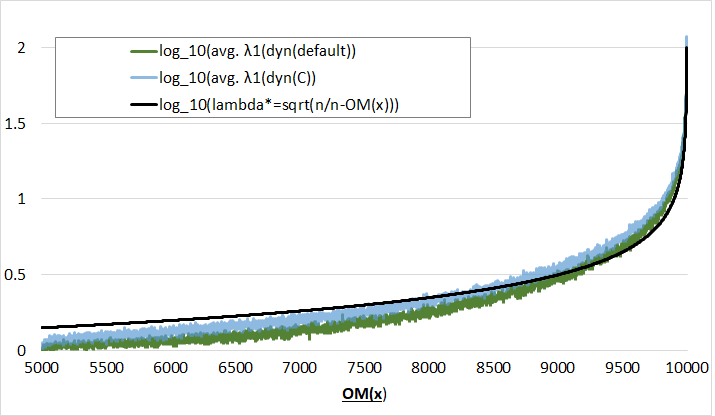}
\caption{Evolution of the average value of $\lambda_1$ over achieved $\OM(x)$ value, for 100 independent runs on the $10\,000$-dimensional \onemax problem, plotted against $\lambda^*=\sqrt{n/(n-\OM(x))}$.}
\label{fig:param-lambda}
\end{figure}
\fi

\section{Tuning the Static \texorpdfstring{\ga}{(1+(l,l)) GA}}%
\label{sec:static}

We had concentrated in the previous sections on optimizing dynamic versions of the \ga, since the theoretical results guarantee configurations for which linear expected running time can be obtained. In contrast, the best possible expected running time that can be achieved with static parameters $\lambda=\lambda_1=\lambda_2$, and arbitrary $p$ and $c$ is of order $n \sqrt{\log(n) \log\log\log(n) / \log\log(n)}$~\cite{DoerrD18ga}. While this rules out the possibility that there exists a static configuration that performs similarly well as $\dyn(C)$ across all dimensions, it is not known to date whether for concrete problem dimensions there exist static configurations that are similar in performance than the dynamic variants $\dyn(\text{default})$, dyn(C), or even $\dyn(\alpha,\beta,\gamma,A,b)$. We next show that for the tested problem dimensions between 500 and $10\,000$ this does not seem to be the case. 

We study the four-dimensional variant $\stat(\lambda_1,\lambda_2,p,c)$ presented in Algorithm~\ref{alg:stat}. Following~\cite{DoerrDE15}, we enforce again that the mutation strength $\ell$ is strictly greater than zero, by sampling from the conditional distribution $\Bin_{>0}(n,p)$ in line~\ref{line:L}. We also allow $\lambda_1 \neq \lambda_2$, which was not the case in~\cite{DoerrDE15}. In line with suggestions from~\cite{DoerrDE15,DoerrD18ga} we set $p=k/n$, and optimize for integer $k \in \{1,\ldots,100\}$. We allow the same range for $\lambda_1$ and $\lambda_2$. The crossover bias $c$ is optimized within the range $[0.01,1/2]$.

\ifgecco
\setlength{\intextsep}{1\baselineskip}
\fi
\begin{algorithm2e}[b]
	\textbf{Initialization:} 
	Choose $x \in \{0,1\}^n$ u.a.r.\;
 \textbf{Optimization:}
\For{$t=1,2,3,\ldots$}{
\underline{\textbf{Mutation phase:}}\\
\Indp
Sample $\ell_1$ from $\Bin_{>0}(n,p=k/n)$\label{line:L}\;
\lFor{$i=1, \ldots, \lambda_1$\label{line:mutstart}}{$x^{(i)} \assign \mut_{\ell}(x)$}
Choose $x' \in \{x^{(1)}, \ldots, x^{(\lambda_1)}\}$ with $f(x')=\max\{f(x^{(1)}), \ldots, f(x^{(\lambda_1)})\}$ u.a.r.\label{line:mutend}\;
\Indm
\underline{\textbf{Crossover phase:}}\\
\Indp
\lFor{$i=1, \ldots, \lambda_2$\label{line:costart}}{$y^{(i)} \assign \cross_{c}(x,x')$}
Choose $y \in \{y^{(1)}, \ldots, y^{(\lambda_2)}\}$ with $f(y)=\max\{f(y^{(1)}), \ldots, f(y^{(\lambda_2)})\}$ u.a.r.\label{line:coend}\;
\Indm
\underline{\textbf{Selection step:}}
\lIf{$f(y)\geq f(x)$}{$x \assign y$
}
}
\caption{The static \ga variant $\stat(\lambda_1,\lambda_2,p=k/n,c)$ with four static parameters.}
\label{alg:stat}
\end{algorithm2e}

\begin{figure}
\centering
\includegraphics[width=0.7\linewidth]{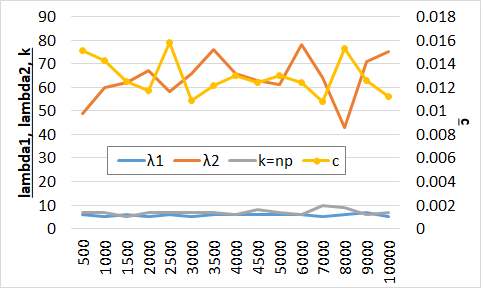}
\caption{Suggested hyper-parameters for the $\stat(\lambda_1,\lambda_2,p=k/n,c)$ by dimension. $\lambda_1$, $\lambda_2$, and $k$
 use the scale on the left, $c$ the one on the right.}
\label{fig:param-stat01}
\ifgecco
\Description{The description of this plot is given in the main text.}
\fi
\end{figure}

\begin{figure*}[t]
\centering
\includegraphics[width=\linewidth]{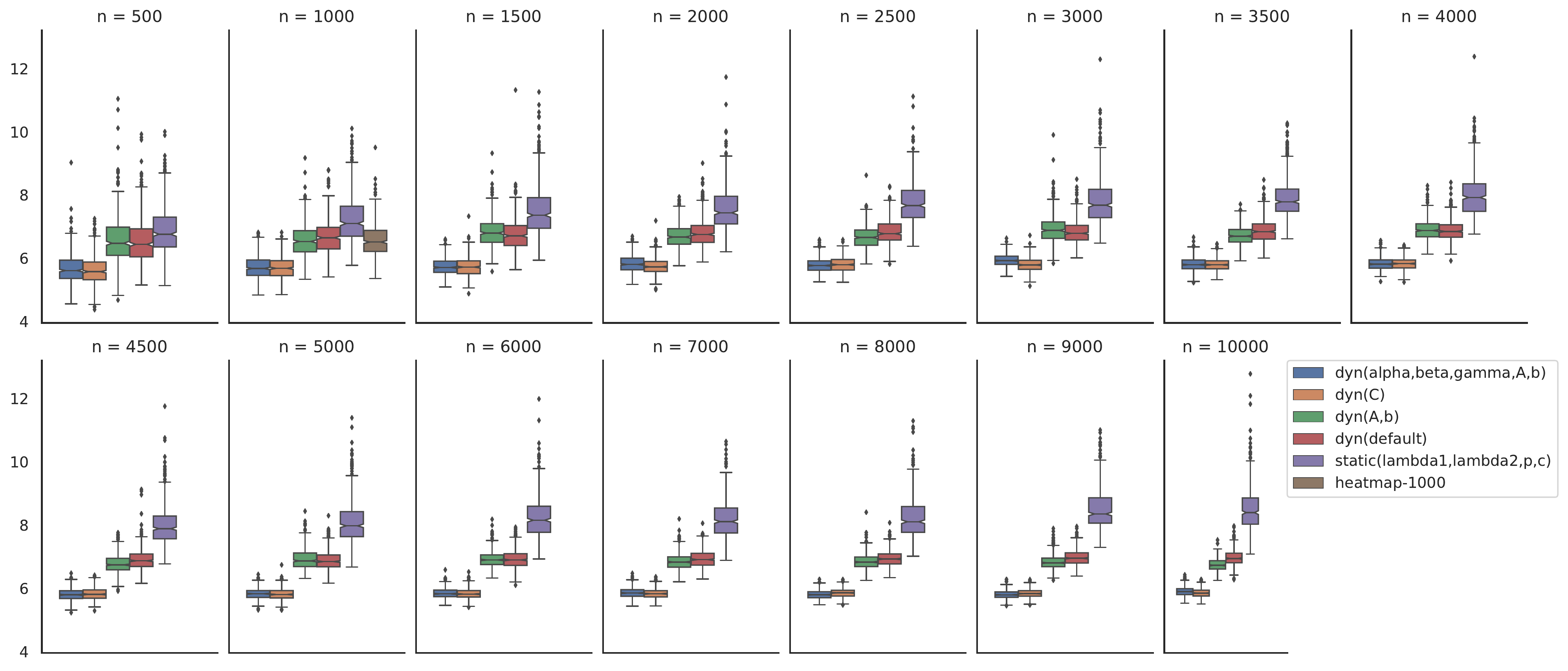}
\caption{Distribution of the by $n$ normalized optimization times of different \ga variants. Heatmap-1000 refers to $\dyn(1,1,1,1.06,0.82)$, which was the best configuration identified in the heatmap from Section~\ref{sec:heatmap}}
\label{fig:boxplot}
\ifgecco
\Description{The description of this plot is given in the main text.}
\fi
\end{figure*}

The normalized average running time of the best configuration that irace has been able to identify with its given budget are reported in column $\stat(\lambda_1,\lambda_2,p=k/n,c)$ of Figure~\ref{fig:runtime-summary}. 
We observe that these running times are significantly larger than those of the dynamic \ga variants. The relative disadvantage against the default dynamic variant $\dyn(\text{default})$ monotonically increases from around $5\%$ for $n=500$ to around $22\%$ for $n=10,000$. Against the best dynamic variant $\dyn(\alpha,\beta,\gamma,A,b)$ this relative disadvantage increases from around $21\%$ to around $44\%$. 

We also see from the results in Figure~\ref{fig:runtime-summary} that, with few exceptions, the normalized average running time increases with the problem dimension. This is in line with what the super-linear lower bound proven in~\cite{DoerrD18ga} suggests (note, however, that the theoretical results for the static \ga assumes $\lambda_1=\lambda_2$). The relative increase of the normalized average running time is smaller than for RLS, again in line with the known theoretical results. The comparison with RLS also shows that the static \ga variants start to outperform RLS at problem dimension $3\,000$. For $n=10\,000$ the relative advantage of $\stat(\lambda_1,\lambda_2,p,c)$ over RLS is around $6\%$. 

Finally, we study in Figure~\ref{fig:param-stat01} the parameter values of the configurations suggested by irace. We observe that across all dimensions $\lambda_1$ is significantly smaller than $\lambda_2$, which was different for the dynamic \ga variants. Both $\lambda_1$ and $k$ are relatively stable, with values ranging between 5 and 7 for $\lambda_1$ and between 5 and 10 for $k$. The values of $\lambda_2$ fluctuates significantly more, between 43 and 78. 
The crossover rate is always within the range $[0.0108,0.0158]$, and thus also quite stable. Since in the original works $c=1/\lambda$ is assumed, we also note that for both $c\lambda_1$ and $c\lambda_2$ the factor between the minimal and maximal value is as small as $1.8$ and $1.5$, respectively, with no clear monotonic relationship. 

\balance
\section{Runtime Distribution}
\label{sec:distribution}\label{sec:boxplots}

In all figures mentioned above we have only considered average values, to obtain results that are more easily comparable with existing theoretical and empirical works. With Figure~\ref{fig:boxplot} we address the question how the running times are distributed. This figure provides boxplots for all tested dimensions $\le 10\,000$. The plots confirm the performance advantages of the five-dimensional dynamic \ga variants $\dyn(\alpha,\beta,\gamma,A,b)$ and dyn(C) over the 2-dimensional versions $\dyn(1,1,1,A,b)$ and $\dyn(\text{default})$. All adaptive versions perform consistently better than the best static version $\stat(\lambda_1,\lambda_2,p,c)$ in term of both median values and variance. These advantages get more visible as the problem sizes increase. We also perform two types of statistical tests - paired Student t-test and Wilcoxon signed-rank test - between those versions. Results confirm that the difference between them are statistically significant with a confidence level of 99.9\%.


\section{Conclusion}
\label{sec:conclusion}

We have presented a very detailed study of the hyper-parameters of the static and the self-adjusting \ga on the \onemax problem. Among other results, we have seen that the self-adjusting \ga gains only around $1\%-3\%$ in average optimization time with optimized update strengths $A$ and $b$. We have then introduced a more flexible variant, the $\dyn(\alpha, \beta,\gamma, A,b)$, in which the offspring population sizes of mutation and crossover phase need not be identical, and which offers more flexibility in the choice of the mutation rate and the crossover bias. This has reduced the average optimization times by another 15\%. Interestingly, the parameter values by which these performance gains are achieved are quite consistent across all tested dimensions. We then analyzed a configuration in which we fixed the hyper-parameters according to the suggestions made by the tuning in lower dimensions $500$ to $10\,000$, and show that it performs very well also on the $20\,000$ and $30\,000$ dimensional \onemax problem. 

Our results suggest that the \ga can gain performance by introducing the additional hyper-parameters. We plan on investigating the gains for other problems, in particular the MaxSAT instances studied in~\cite{BuzdalovD17}. Since all results shown in this work are quite consistent across all dimensions, we also plan on analyzing the advantages of the $\dyn(\alpha,\beta,\gamma, A,b)$ by rigorous means, both in terms of optimization time, but also in terms of more general fixed-target running times. As we have demonstrated in Section~\ref{sec:FT}, the latter reveal that the advantage of the \ga over RLS lies in the very final phases of the \onemax optimization problem, i.e., when finding improving moves is hard. Efficiently switching between the two algorithms at the time at which the \ga starts to outperform RLS carries the potential to reduce the optimization time further. 
Automating such online algorithm selection is another line of research that we plan to investigate further. Techniques from the literature on parameter control~\cite{KarafotiasHE15,DoerrD18chapter}, adaptive operator selection~\cite{FialhoCSS10}, and hyper-heuristics~\cite{BurkeGHKOOQ13} might prove useful in this context.  

\ifarxiv
On a meta-level, we have demonstrated with this work that hyper-parameter tuning provides useful insights that help us understand the working principles of randomized search heuristics. 
As mentioned, we are confident that we can leverage the empirical findings of this work for a precise theoretical analysis of the self-adjusting \ga. Similar to the \emph{tuning in the loop} approach suggested in~\cite{loop}, we thus see that there is important room for \emph{tuning in the theory loop}. 
On the other hand, we also show an example that raises the question of how to adjust the default setting of current hyper-parameter tuning methods when a priori knowledge about the scenario is given (e.g., high variance in performance measure in our case). This is another line of research raised by this work. 
\fi

\subsubsection*{\textbf{Acknowledgments.}}

This work was supported by the Paris Ile-de-France Region, by a public grant as part of the Investissement d'avenir project ANR-11-LABX-0056-LMH, LabEx LMH, by the European Cooperation in Science and Technology (COST) action CA15140, and by the UK EPSRC grant EP/P015638/1. 
The simulations were performed at the HPCaVe at UPMC-Sorbonne Universit\'e.

}


\begin{thebibliography}{HHLBS09}

\bibitem[aFU]{PyImp}
Ml4AAD~Group at~Freiburg~University.
\newblock Pyimp.
\newblock \url{https://github.com/automl/ParameterImportance}.

\bibitem[AM16]{AletiM16}
Aldeida Aleti and Irene Moser.
\newblock A systematic literature review of adaptive parameter control methods
  for evolutionary algorithms.
\newblock {\em ACM Computing Surveys}, 49:56:1--56:35, 2016.

\bibitem[AMS{\etalchar{+}}15]{GGA}
Carlos Ans\'{o}tegui, Yuri Malitsky, Horst Samulowitz, Meinolf Sellmann, and
  Kevin Tierney.
\newblock Model-based genetic algorithms for algorithm configuration.
\newblock In {\em Proc. of International Conference on Artificial Intelligence
  (IJCAI'15)}, pages 733--739. AAAI Press, 2015.

\bibitem[aSU]{HPCaVe}
HPCaVe~Cluster at~Sorbonne~University.
\newblock \url{http://hpcave.upmc.fr/index.php/resources/mesu-beta/}.

\bibitem[Aug09]{Auger09}
Anne Auger.
\newblock Benchmarking the (1+1) evolution strategy with one-fifth success rule
  on the {BBOB}-2009 function testbed.
\newblock In {\em Companion Material for Proc. of Genetic and Evolutionary
  Computation Conference (GECCO'09)}, pages 2447--2452. ACM, 2009.

\bibitem[BD17]{BuzdalovD17}
Maxim Buzdalov and Benjamin Doerr.
\newblock Runtime analysis of the $(1+(\lambda,\lambda))$ {Genetic Algorithm}
  on random satisfiable {3-CNF} formulas.
\newblock In {\em Proc. of Genetic and Evolutionary Computation Conference
  (GECCO'17)}, pages 1343--1350. ACM, 2017.

\bibitem[BGH{\etalchar{+}}13]{BurkeGHKOOQ13}
Edmund~K. Burke, Michel Gendreau, Matthew~R. Hyde, Graham Kendall, Gabriela
  Ochoa, Ender {\"{O}}zcan, and Rong Qu.
\newblock Hyper-heuristics: a survey of the state of the art.
\newblock {\em Journal of the Operational Research Society}, 64:1695--1724,
  2013.

\bibitem[CD18]{CarvalhoD18}
Eduardo {Carvalho Pinto} and Carola Doerr.
\newblock A simple proof for the usefulness of crossover in black-box
  optimization.
\newblock In {\em Proc. of Parallel Problem Solving from Nature (PPSN'18)},
  volume 11102 of {\em Lecture Notes in Computer Science}, pages 29--41.
  Springer, 2018.
\newblock Full version available at \url{http://arxiv.org/abs/1812.00493}.

\bibitem[CLHS17]{iracecapping}
Leslie~P{\'{e}}rez C{\'{a}}ceres, Manuel L{\'{o}}pez{-}Ib{\'{a}}{\~{n}}ez,
  Holger Hoos, and Thomas St{\"{u}}tzle.
\newblock An experimental study of adaptive capping in irace.
\newblock In {\em Proc. of Learning and Intelligent Optimization (LION'17)},
  volume 10556 of {\em Lecture Notes in Computer Science}, pages 235--250.
  Springer, 2017.

\bibitem[DD18a]{DoerrD18ga}
Benjamin Doerr and Carola Doerr.
\newblock Optimal static and self-adjusting parameter choices for the
  $(1+(\lambda,\lambda))$ genetic algorithm.
\newblock {\em Algorithmica}, 80:1658--1709, 2018.

\bibitem[DD18b]{DoerrD18chapter}
Benjamin Doerr and Carola Doerr.
\newblock Theory of parameter control mechanisms for discrete black-box
  optimization: Provable performance gains through dynamic parameter choices.
\newblock In Benjamin Doerr and Frank Neumann, editors, {\em Theory of
  Randomized Search Heuristics in Discrete Search Spaces}. Springer, 2018.
\newblock To appear. Available online at
  \url{https://arxiv.org/abs/1804.05650}.

\bibitem[DD19]{Supplementary}
Nguyen Dang and Carola Doerr.
\newblock Github repository with project data and plots.
\newblock \url{https://github.com/ndangtt/1LLGA}, 2019.

\bibitem[DDE15]{DoerrDE15}
Benjamin Doerr, Carola Doerr, and Franziska Ebel.
\newblock From black-box complexity to designing new genetic algorithms.
\newblock {\em Theoretical Computer Science}, 567:87--104, 2015.

\bibitem[DDY16]{DoerrDY16}
Benjamin Doerr, Carola Doerr, and Jing Yang.
\newblock Optimal parameter choices via precise black-box analysis.
\newblock In {\em Proc. of Genetic and Evolutionary Computation Conference
  (GECCO'16)}, pages 1123--1130. {ACM}, 2016.

\bibitem[DJS{\etalchar{+}}13]{DoerrJSWZ13}
Benjamin Doerr, Thomas Jansen, Dirk Sudholt, Carola Winzen, and Christine
  Zarges.
\newblock Mutation rate matters even when optimizing monotonic functions.
\newblock {\em Evolutionary Computation}, 21:1--27, 2013.

\bibitem[dOAS11]{loop}
Marco Antonio~Montes de~Oca, Dogan Aydin, and Thomas St{\"{u}}tzle.
\newblock An incremental particle swarm for large-scale continuous optimization
  problems: an example of tuning-in-the-loop (re)design of optimization
  algorithms.
\newblock {\em Soft Comput.}, 15:2233--2255, 2011.

\bibitem[DWY{\etalchar{+}}18]{IOHprofiler}
Carola Doerr, Hao Wang, Furong Ye, Sander van Rijn, and Thomas B{\"{a}}ck.
\newblock {IOHprofiler:} {A} benchmarking and profiling tool for iterative
  optimization heuristics.
\newblock {\em CoRR}, abs/1810.05281, 2018.
\newblock IOHprofiler is available at \url{https://github.com/IOHprofiler}.

\bibitem[FCSS10]{FialhoCSS10}
{\'{A}}lvaro Fialho, Lu{\'{\i}}s~Da Costa, Marc Schoenauer, and Mich{\`{e}}le
  Sebag.
\newblock Analyzing bandit-based adaptive operator selection mechanisms.
\newblock {\em Annals of Mathematics and Artificial Intelligence}, 60:25--64,
  2010.

\bibitem[HHLB11]{SMAC}
Frank Hutter, Holger~H. Hoos, and Kevin Leyton-Brown.
\newblock Sequential model-based optimization for general algorithm
  configuration.
\newblock In {\em Proc. of Learning and Intelligent Optimization (LION'11)},
  pages 507--523. Springer, 2011.

\bibitem[HHLBS09]{ParamILS}
Frank Hutter, Holger~H. Hoos, Kevin Leyton-Brown, and Thomas St\"{u}tzle.
\newblock Param{ILS}: An automatic algorithm configuration framework.
\newblock {\em Journal of Artificial Intelligence Research}, 36:267--306, 2009.

\bibitem[HLB14]{hoos2014efficient}
Holger Hoos and Kevin Leyton-Brown.
\newblock An efficient approach for assessing hyperparameter importance.
\newblock In {\em International Conference on Machine Learning}, pages
  754--762, 2014.

\bibitem[KHE15]{KarafotiasHE15}
Giorgos Karafotias, Mark Hoogendoorn, and A.E. Eiben.
\newblock Parameter control in evolutionary algorithms: Trends and challenges.
\newblock {\em IEEE Transactions on Evolutionary Computation}, 19:167--187,
  2015.

\bibitem[KMH{\etalchar{+}}04]{KernMHBOK04}
Stefan Kern, Sibylle~D. M{\"{u}}ller, Nikolaus Hansen, Dirk B{\"{u}}che, Jiri
  Ocenasek, and Petros Koumoutsakos.
\newblock Learning probability distributions in continuous evolutionary
  algorithms - a comparative review.
\newblock {\em Natural Computing}, 3:77--112, 2004.

\bibitem[LDC{\etalchar{+}}16]{irace}
Manuel L{\'{o}}pez{-}Ib{\'{a}}{\~{n}}ez, J{\'{e}}r{\'{e}}mie Dubois{-}Lacoste,
  Leslie~P{\'{e}}rez C{\'{a}}ceres, Mauro Birattari, and Thomas St{\"{u}}tzle.
\newblock The irace package: Iterated racing for automatic algorithm
  configuration.
\newblock {\em Operations Research Perspectives}, 3:43--58, 2016.

\bibitem[Len18]{Lengler18}
Johannes Lengler.
\newblock A general dichotomy of evolutionary algorithms on monotone functions.
\newblock In {\em Proc. of Parallel Problem Solving from Nature (PPSN'18)},
  volume 11102 of {\em Lecture Notes in Computer Science}, pages 3--15.
  Springer, 2018.

\bibitem[LJD{\etalchar{+}}17]{Hyperband}
Lisha Li, Kevin~G. Jamieson, Giulia DeSalvo, Afshin Rostamizadeh, and Ameet
  Talwalkar.
\newblock Hyperband: {A} novel bandit-based approach to hyperparameter
  optimization.
\newblock {\em Journal of Machine Learning Research}, 18:185:1--185:52, 2017.

\bibitem[LW12]{LehreW12}
Per~Kristian Lehre and Carsten Witt.
\newblock Black-box search by unbiased variation.
\newblock {\em Algorithmica}, 64:623--642, 2012.

\end{thebibliography}
\newcommand{\etalchar}[1]{$^{#1}$}

\ifgecco
\onecolumn
\fi

\ifarxiv
\newpage
\appendix
\section{Selected Optimization Time Statistics}

\begin{table}[h]
\begin{tabular}{rrrrrrrrrrr}
& \multicolumn{5}{c}{quantiles}&&&&&\\
\cmidrule(lr){2-6}
$n$     & 20\%   & 25\%   & 50\%   & 75\%   & 98\%   & mean   & rsd & $A$    & $b$    & success rate \\
\hline
500   & 3,009  & 3,048  & 3,239  & 3,493  & 4,214  & 3,296  & 11.5        & 1.11 & 0.65 & 5.05         \\
1,000  & 6,118  & 6,207  & 6,534  & 6,876  & 7,719  & 6,573  & 7.9         & 1.07 & 0.79 & 4.52         \\
1,500  & 9,682  & 9,764  & 10,199 & 10,646 & 11,775 & 10,267 & 6.6         & 1.08 & 0.63 & 6.68         \\
2,000  & 12,764 & 12,897 & 13,349 & 13,877 & 15,082 & 13,411 & 5.6         & 1.08 & 0.73 & 5.29         \\
2,500  & 15,879 & 16,038 & 16,650 & 17,241 & 18,649 & 16,683 & 5.5         & 1.05 & 0.85 & 4.11         \\
3,000  & 19,733 & 19,896 & 20,653 & 21,458 & 23,924 & 20,778 & 6.2         & 1.12 & 0.63 & 4.92         \\
3,500  & 22,675 & 22,808 & 23,458 & 24,196 & 25,902 & 23,537 & 4.3         & 1.06 & 0.79 & 5.02         \\
4,000  & 26,573 & 26,730 & 27,518 & 28,378 & 30,688 & 27,639 & 4.7         & 1.11 & 0.64 & 5.40         \\
4,500  & 29,368 & 29,649 & 30,354 & 31,289 & 33,506 & 30,494 & 4.3         & 1.07 & 0.78 & 4.71         \\
5,000  & 33,243 & 33,454 & 34,358 & 35,607 & 38,232 & 34,601 & 4.6         & 1.10 & 0.63 & 5.74         \\
6,000  & 40,279 & 40,543 & 41,406 & 42,373 & 45,166 & 41,535 & 3.7         & 1.09 & 0.65 & 6.01         \\
7,000  & 46,469 & 46,712 & 47,840 & 48,992 & 52,307 & 47,977 & 3.8         & 1.08 & 0.74 & 5.01         \\
8,000  & 53,206 & 53,529 & 54,666 & 55,956 & 58,903 & 54,807 & 3.5         & 1.07 & 0.76 & 5.08         \\
9,000  & 59,949 & 60,206 & 61,279 & 62,618 & 66,868 & 61,547 & 3.4         & 1.07 & 0.76 & 4.96         \\
10,000 & 65,761 & 66,144 & 67,315 & 68,767 & 71,414 & 67,444 & 2.8         & 1.04 & 0.86 & 4.89       \\ 
\vspace{0.25ex}
\end{tabular}
\caption{Performance statistics for 500 independent runs of the configurations suggested by irace for $\dyn(1,1,1,A,b)$. rsd abbreviates ``relative standard deviation'', i.e., standard deviation divided by the mean value.}
\label{tab:dyn01-stats}
\end{table}

\begin{table}[h]
\small
\setlength\tabcolsep{2pt}
\begin{tabular}{rrrrrrrrrrrrrr}
& \multicolumn{5}{c}{quantiles}&&&&&\\
\cmidrule(lr){2-6}
$n$     & 20\%   & 25\%   & 50\%   & 75\%   & 98\%   & mean   & rsd & $\alpha$ & $\beta$  & $\gamma$ & $A$     & $b$     & success rate \\
\hline
500   & 2,659  & 2,683  & 2,807  & 2,972  & 3,328  & 2,835  & 8.1 & 0.522 & 1.854 & 0.876 & 1.229 & 0.598 & 3.497        \\
1,000  & 5,414  & 5,462  & 5,680  & 5,949  & 6,447  & 5,711  & 6.1 & 0.428 & 1.438 & 1.053 & 1.157 & 0.740 & 3.063        \\
1,500  & 8,281  & 8,336  & 8,568  & 8,870  & 9,365  & 8,600  & 4.6 & 0.378 & 1.727 & 1.004 & 1.143 & 0.736 & 3.288        \\
2,000  & 11,193 & 11,278 & 11,616 & 12,009 & 12,795 & 11,652 & 4.5 & 0.414 & 1.380 & 1.125 & 1.153 & 0.660 & 3.913        \\
2,500  & 13,973 & 14,064 & 14,432 & 14,801 & 15,865 & 14,472 & 4.1 & 0.473 & 1.494 & 1.150 & 1.145 & 0.723 & 3.391        \\
3,000  & 17,333 & 17,428 & 17,782 & 18,206 & 19,065 & 17,822 & 3.2 & 0.504 & 2.524 & 0.619 & 1.255 & 0.526 & 3.824        \\
3,500  & 19,702 & 19,855 & 20,296 & 20,822 & 21,861 & 20,336 & 3.6 & 0.441 & 1.686 & 0.842 & 1.160 & 0.702 & 3.386        \\
4,000  & 22,679 & 22,762 & 23,262 & 23,811 & 25,133 & 23,325 & 3.3 & 0.426 & 1.720 & 0.896 & 1.168 & 0.675 & 3.539        \\
4,500  & 25,473 & 25,566 & 26,095 & 26,676 & 27,788 & 26,133 & 3.1 & 0.363 & 1.429 & 1.202 & 1.149 & 0.719 & 3.372        \\
5,000  & 28,454 & 28,572 & 29,162 & 29,670 & 31,114 & 29,165 & 2.9 & 0.359 & 1.413 & 1.238 & 1.167 & 0.691 & 3.391        \\
6,000  & 34,238 & 34,436 & 34,978 & 35,694 & 37,000 & 35,056 & 2.7 & 0.373 & 1.654 & 1.070 & 1.164 & 0.711 & 3.255        \\
7,000  & 40,065 & 40,260 & 40,982 & 41,733 & 43,422 & 41,021 & 2.7 & 0.342 & 1.187 & 1.227 & 1.109 & 0.738 & 3.934        \\
8,000  & 45,477 & 45,660 & 46,410 & 47,178 & 48,546 & 46,412 & 2.3 & 0.490 & 1.606 & 0.954 & 1.110 & 0.783 & 3.352        \\
9,000  & 51,284 & 51,464 & 52,176 & 52,995 & 54,696 & 52,248 & 2.3 & 0.447 & 1.447 & 1.109 & 1.106 & 0.779 & 3.482        \\
10,000 & 57,852 & 58,064 & 59,013 & 59,931 & 62,026 & 59,033 & 2.4 & 0.435 & 1.271 & 1.111 & 1.141 & 0.722 & 3.475       \\
\vspace{0.25ex}
\end{tabular}
\caption{Performance statistics for 500 independent runs of the configurations suggested by irace for $\dyn(\alpha,\beta,\gamma,A,b)$. rsd abbreviates ``relative standard deviation'', i.e., standard deviation divided by the mean value.}
\label{tab:dyn02-stats}
\end{table}

\begin{table}[h]
\begin{tabular}{rrrrrrrrrrrr}
& \multicolumn{5}{c}{quantiles}&&&&&\\
\cmidrule(lr){2-6}
$n$   & 20\%   & 25\%   & 50\%   & 75\%   & 98\%    & mean   & rsd  & $\lambda_1$ & $\lambda_2$ & $k$ & $c$    \\
\hline
500   & 3,142  & 3,181  & 3,382  & 3,652  & 4,386   & 3,437  & 11.2 & 6           & 49          & 7   & 0.0151 \\
1,000  & 6,599  & 6,702  & 7,102  & 7,650  & 9,124   & 7,225  & 10.3 & 5           & 60          & 7   & 0.0143 \\
1,500  & 10,321 & 10,428 & 11,048 & 11,880 & 14,492  & 11,277 & 10.7 & 6           & 62          & 5   & 0.0125 \\
2,000  & 13,951 & 14,178 & 14,884 & 15,930 & 18,409  & 15,130 & 9.6  & 5           & 67          & 7   & 0.0117 \\
2,500  & 18,056 & 18,228 & 19,178 & 20,376 & 23,125  & 19,398 & 8.6  & 6           & 58          & 7   & 0.0158 \\
3,000  & 21,545 & 21,867 & 23,049 & 24,551 & 29,181  & 23,373 & 9.5  & 5           & 66          & 7   & 0.0109 \\
3,500  & 25,946 & 26,218 & 27,258 & 28,670 & 33,538  & 27,677 & 7.9  & 6           & 76          & 7   & 0.0121 \\
4,000  & 29,619 & 29,950 & 31,698 & 33,432 & 39,096  & 32,034 & 9    & 6           & 66          & 6   & 0.013  \\
4,500  & 33,727 & 34,072 & 35,502 & 37,298 & 43,695  & 36,049 & 8.2  & 6           & 63          & 8   & 0.0124 \\
5,000  & 37,728 & 38,200 & 39,900 & 42,145 & 49,722  & 40,540 & 8.6  & 6           & 61          & 7   & 0.013  \\
6,000  & 46,126 & 46,656 & 48,946 & 51,592 & 58,755  & 49,475 & 8    & 6           & 78          & 6   & 0.0124 \\
7,000  & 53,793 & 54,275 & 56,802 & 59,802 & 67,465  & 57,401 & 7.5  & 5           & 64          & 10  & 0.0108 \\
8,000  & 61,621 & 62,201 & 64,872 & 68,728 & 79,383  & 65,957 & 8.1  & 6           & 43          & 9   & 0.0153 \\
9,000  & 71,811 & 72,596 & 75,198 & 79,792 & 92,431  & 76,500 & 7.8  & 7           & 71          & 6   & 0.0126 \\
10,000 & 79,349 & 80,324 & 83,984 & 88,606 & 102,910 & 85,087 & 8.6  & 5           & 75          & 7   & 0.0112 \\
\vspace{0.25ex}
\end{tabular}
\caption{Performance statistics for 500 independent runs of the configurations suggested by irace for $\stat(\lambda_1,\lambda_2,k,c)$. rsd abbreviates ``relative standard deviation'', i.e., standard deviation divided by the mean value.}
\label{tab:stat01-stats}
\end{table}

\begin{table}[h]
\small
\setlength\tabcolsep{2pt}
\begin{tabular}{rrrrrrrrrr}
& \multicolumn{5}{c}{quantiles}& & &\\
\cmidrule(lr){3-7}
configuration & $n$     & 20\%   & 25\%   & 50\%   & 75\%   & 98\%   & mean   & rsd & mean/$n$\\
\hline
dyn(C)        & 500   & 2626.8   & 2662.75   & 2789.5   & 2941.5    & 3361.82   & 2810.562   & 8.32        & 5.62                \\
dyn(C2)       & 500   & 2714     & 2747      & 2894     & 3033      & 3392.06   & 2904.076   & 7.69        & 5.81                \\
dyn(default)  & 500   & 2981     & 3024.75   & 3222     & 3467.25   & 4166.72   & 3278.202   & 11.04       & 6.56                \\
\hline
dyn(C)        & 1000  & 5417.8   & 5454.5    & 5686     & 5926.75   & 6398.26   & 5695.144   & 5.76        & 5.70                \\
dyn(C2)       & 1000  & 5580.8   & 5645.75   & 5869     & 6112.25   & 6668.9    & 5893.592   & 6.04        & 5.89                \\
dyn(default)  & 1000  & 6232     & 6299.5    & 6652     & 6980.25   & 7847.54   & 6671.378   & 7.86        & 6.67                \\
\hline
dyn(C)        & 1500  & 8203.4   & 8277.75   & 8586     & 8880.75   & 9528.1    & 8591.36    & 5.21        & 5.73                \\
dyn(C2)       & 1500  & 8546     & 8603.5    & 8928     & 9206.75   & 9868.02   & 8919.98    & 4.96        & 5.95                \\
dyn(default)  & 1500  & 9528     & 9610.75   & 10068    & 10554.75  & 11866.46  & 10164.5    & 7.55        & 6.78                \\
\hline
dyn(C)        & 2000  & 11099.4  & 11172     & 11465    & 11805     & 12716.14  & 11495.294  & 4.66        & 5.75                \\
dyn(C2)       & 2000  & 11450.8  & 11532.5   & 11881    & 12223.5   & 12951.24  & 11893.274  & 4.21        & 5.95                \\
dyn(default)  & 2000  & 12919.4  & 13010.75  & 13514    & 14079.25  & 15930.84  & 13636.004  & 6.9         & 6.82                \\
\hline
dyn(C)        & 2500  & 13985.8  & 14076.25  & 14502    & 14910     & 15772.26  & 14509.296  & 4.1         & 5.80                \\
dyn(C2)       & 2500  & 14431.6  & 14556.75  & 14903.5  & 15330.5   & 16292.36  & 14942.674  & 4.01        & 5.98                \\
dyn(default)  & 2500  & 16342.4  & 16452.75  & 16948    & 17717.75  & 19403.6   & 17071.06   & 5.78        & 6.83                \\
\hline
dyn(C)        & 3000  & 16861    & 16962.25  & 17376.5  & 17821     & 18880.14  & 17408.882  & 3.7         & 5.80                \\
dyn(C2)       & 3000  & 17486.2  & 17582     & 17997    & 18444.5   & 19379.24  & 18021.416  & 3.6         & 6.01                \\
dyn(default)  & 3000  & 19606    & 19750.75  & 20374    & 21125.25  & 23078.9   & 20492.278  & 5.4         & 6.83                \\
\hline
dyn(C)        & 3500  & 19759    & 19844.5   & 20286.5  & 20736.75  & 21958.32  & 20329.532  & 3.34        & 5.81                \\
dyn(C2)       & 3500  & 20526.2  & 20650     & 21029.5  & 21518.25  & 22560.68  & 21080.554  & 3.35        & 6.02                \\
dyn(default)  & 3500  & 22966.2  & 23138.75  & 23958    & 24812.5   & 27550.58  & 24090.904  & 5.4         & 6.88                \\
\hline
dyn(C)        & 4000  & 22674.8  & 22785.5   & 23333.5  & 23799.5   & 24974.34  & 23300.608  & 3.24        & 5.83                \\
dyn(C2)       & 4000  & 23394.8  & 23511.5   & 24028    & 24476.75  & 25979.2   & 24031.834  & 3.32        & 6.01                \\
dyn(default)  & 4000  & 26475.6  & 26681.5   & 27404.5  & 28250.75  & 30781.46  & 27543.488  & 4.81        & 6.89                \\
\hline
dyn(C)        & 4500  & 25492.2  & 25614.25  & 26165.5  & 26790.25  & 28047.02  & 26225.81   & 3.21        & 5.83                \\
dyn(C2)       & 4500  & 26358.2  & 26465.25  & 27053.5  & 27593.25  & 28873.2   & 27039.558  & 3.09        & 6.01                \\
dyn(default)  & 4500  & 29849.8  & 30119.5   & 30932    & 31905.75  & 34210.14  & 31120.334  & 5.06        & 6.92                \\
\hline
dyn(C)        & 5000  & 28394.6  & 28506     & 29045.5  & 29662.25  & 31293.42  & 29118.248  & 3.14        & 5.82                \\
dyn(C2)       & 5000  & 29372    & 29533.75  & 30082.5  & 30720     & 32104.12  & 30148.46   & 3.01        & 6.03                \\
dyn(default)  & 5000  & 33250.6  & 33419.75  & 34246.5  & 35297.25  & 38290.32  & 34445.674  & 4.43        & 6.89                \\
\hline
dyn(C)        & 6000  & 34199.8  & 34371.75  & 34931    & 35606.5   & 37094.12  & 35010.002  & 2.71        & 5.84                \\
dyn(C2)       & 6000  & 35347.6  & 35558.75  & 36206    & 36879.25  & 38274.04  & 36215.806  & 2.64        & 6.04                \\
dyn(default)  & 6000  & 40115.2  & 40376.25  & 41403    & 42590     & 46005.26  & 41583.822  & 4.34        & 6.93                \\
\hline
dyn(C)        & 7000  & 39937.8  & 40122.5   & 40832    & 41505.75  & 43196.24  & 40855.698  & 2.6         & 5.84                \\
dyn(C2)       & 7000  & 41354.4  & 41556.75  & 42195    & 42814.75  & 44653.86  & 42235.238  & 2.48        & 6.03                \\
dyn(default)  & 7000  & 46882.8  & 47156.25  & 48378.5  & 49780.5   & 53550.36  & 48597.422  & 4.11        & 6.94                \\
\hline
dyn(C)        & 8000  & 45862    & 46084     & 46886.5  & 47539     & 49443.74  & 46865.952  & 2.48        & 5.86                \\
dyn(C2)       & 8000  & 47440.8  & 47651.75  & 48352.5  & 49104.5   & 51046.26  & 48391.546  & 2.44        & 6.05                \\
dyn(default)  & 8000  & 53844.8  & 54192     & 55451    & 56728     & 60514.12  & 55551.484  & 3.69        & 6.94                \\
\hline
dyn(C)        & 9000  & 51605.6  & 51799.75  & 52525.5  & 53357.75  & 55259.4   & 52609.86   & 2.33        & 5.85                \\
dyn(C2)       & 9000  & 53401    & 53591.5   & 54402.5  & 55290.5   & 56978.2   & 54434.674  & 2.26        & 6.05                \\
dyn(default)  & 9000  & 60961    & 61244.5   & 62592.5  & 64145.75  & 67863.7   & 62824.338  & 3.52        & 6.98                \\
\hline
dyn(C)        & 10000 & 57456.6  & 57635.25  & 58522    & 59468     & 61273.56  & 58563.928  & 2.19        & 5.86                \\
dyn(C2)       & 10000 & 59399    & 59634.25  & 60561    & 61362.75  & 63161.1   & 60536.716  & 2.12        & 6.05                \\
dyn(default)  & 10000 & 67604.2  & 68112     & 69470.5  & 71157.75  & 74865.24  & 69679.228  & 3.41        & 6.97                \\
\hline
dyn(C)        & 20000 & 116124.4 & 116421.5  & 117812.5 & 119032.25 & 121907.32 & 117795.328 & 1.72        & 5.89                \\
dyn(C2)       & 20000 & 119667.6 & 119940.25 & 121234   & 122663    & 125482.96 & 121326.13  & 1.65        & 6.07                \\
dyn(default)  & 20000 & 137199   & 137519.5  & 139915.5 & 142270.25 & 149804.72 & 140264.274 & 2.77        & 7.01                \\
\hline
dyn(C)        & 30000 & 174611.8 & 175053.75 & 176968.5 & 178318    & 181736    & 176833.916 & 1.39        & 5.89                \\
dyn(C2)       & 30000 & 180407.8 & 180792    & 182259   & 183875.75 & 186964.2  & 182249.43  & 1.3         & 6.07                \\
dyn(default)  & 30000 & 207066.8 & 207579.75 & 210926.5 & 214517.25 & 224366.5  & 211485.7   & 2.57        & 7.05               \\
\vspace{0.25ex}
\end{tabular}
\caption{Performance statistics for 500 independent runs of the self-adjusting \ga with the following configurations: dyn(C)=$\dyn(0.45,1.6,1,1.16,0.7)$, dyn(C2)=$\dyn(1/2,2,1/2,(3/2)^{1/4},2/3)$, and dyn(default)=$\dyn(1,1,1,(3/2)^{1/4},2/3)$. rsd abbreviates ``relative standard deviation'', i.e., standard deviation divided by the mean value.}
\label{tab:default-stats}
\end{table}

\fi

\end{document}